\documentclass[11pt,a4paper]{article}

\usepackage[T1]{fontenc}
\usepackage[utf8]{inputenc}
\usepackage{lmodern}

\usepackage{amsmath,amssymb}
\usepackage{graphicx}
\usepackage{booktabs}
\usepackage{float}
\usepackage{longtable}
\usepackage{tabularx}
\usepackage{array}
\usepackage{ragged2e}

\usepackage{tikz}
\usetikzlibrary{shapes.geometric, positioning, arrows.meta}
\definecolor{linegray}{gray}{0.35}
\definecolor{fillgray}{gray}{0.98}
\definecolor{textgray}{gray}{0.10}

\usepackage[numbers,sort&compress]{natbib}
\usepackage{caption}
\usepackage[hidelinks]{hyperref}

\newcolumntype{L}[1]{>{\RaggedRight\arraybackslash}p{#1}}
\newcolumntype{C}[1]{>{\centering\arraybackslash}p{#1}}

\title{Concerning Uncertainty\\
\large A Systematic Survey of Uncertainty-Aware XAI}

\author{
H.~L\"ofstr\"om$^{1}$,
T.~L\"ofstr\"om$^{2}$,
A.~Hjort$^{3}$,
F.~R.~Yapicioglu$^{4,5}$
}

\date{}

\begin{document}

\maketitle

\begin{center}
\small
$^{1}$ J\"onk\"oping International Business School, Gjuterigatan 5, 553 18 J\"onk\"oping, Sweden\\
$^{2}$ J\"onk\"oping School of Engineering, Gjuterigatan 5, 553 18 J\"onk\"oping, Sweden\\
$^{3}$ Eiendomsverdi AS, Nedre Slottsgate 8, 0157 Oslo, Norway\\
$^{4}$ DISI, University of Bologna, Mura Anteo Zamboni 7, 40126 Bologna, Italy\\
$^{5}$ Marketing and Sales, Automobili Lamborghini S.p.A., Via Modena 12, 40019 Sant'Agata Bolognese, Italy\\[0.5em]
Corresponding author: Helena L\"ofstr\"om (\texttt{helena.lofstrom@ju.se})
\end{center}

\begin{abstract}
\sloppy
This paper surveys  uncertainty-aware explainable artificial intelligence (UAXAI), examining how uncertainty is incorporated into explanatory pipelines and how such methods are evaluated. Across the literature, three recurring approaches to uncertainty quantification emerge (Bayesian,\ Monte Carlo,\ and\ Conformal methods), alongside distinct strategies for integrating uncertainty into explanations: assessing trustworthiness, constraining models or explanations, and explicitly communicating uncertainty. Evaluation practices remain fragmented and largely model centred, with limited attention to users and inconsistent reporting of reliability properties (e.g., calibration, coverage, explanation stability). Recent work leans towards calibration, distribution free techniques and recognises explainer variability as a central concern. We argue that progress in UAXAI requires unified evaluation principles that link uncertainty propagation, robustness, and human decision-making, and highlight counterfactual and calibration approaches as promising avenues for aligning interpretability with reliability.
\end{abstract}

\maketitle

\section{Introduction} \label{introduction}
Artificial intelligence (AI) has rapidly evolved from a promising computational paradigm into a pervasive enabler of innovation across diverse domains such as healthcare and autonomous vehicles. As AI systems increasingly engage in high-stakes decision-making and grow in complexity, their black-box nature raises fundamental concerns regarding transparency, reliability, and trust \cite{mersha2024explainable,xie2023joint,shaheema2025explainable}. A central response to these concerns is provided by explainable artificial intelligence (XAI), which seeks to make model outcomes more interpretable, thereby enabling humans to better understand, trust, and effectively interact with AI systems \cite{forster2025taxonomy}. However, explanations alone are often insufficient if they do not communicate how reliable a model’s prediction or its explanation is, particularly in settings where users must assess confidence and risk.

This survey focuses on uncertainty-aware explainable artificial intelligence (UAXAI), an emerging class of methods that explicitly incorporate uncertainty into the explanatory process; a precise operational definition is formalized later in the paper. In UAXAI, uncertainty is not treated solely as a property of model predictions, but as an integral component of explanations themselves, with the goal of improving transparency, reliability, and appropriate human trust in AI-assisted decision-making. From this perspective, existing UAXAI methods can be broadly understood according to three complementary roles played by uncertainty within explanatory pipelines: (i) approaches that treat uncertainty itself as an explanatory signal, conveying confidence or risk alongside predictions; (ii) approaches that use uncertainty to qualify, constrain, or modulate explanations generated by existing XAI techniques; and (iii) approaches that employ uncertainty quantification to assess the robustness, stability, or reliability of explanations. These roles provide a conceptual framework for organizing the literature reviewed in this survey.

To contextualize how uncertainty is integrated into explanations, we first review state-of-the-art approaches to XAI, which span a wide methodological spectrum. Post-hoc techniques include both factual explanations, which clarify why a model produced a particular output, and counterfactual explanations, which illustrate what minimal change to input features would have resulted in a different outcome. Counterfactual explanations address the fundamental question “Why not another?” Building on these perspectives, post-hoc methods such as LIME \cite{ribeiro2016should}, SHAP \cite{lundberg2017unified}, and counterfactual reasoning frameworks \cite{schut2021generating} aim to rationalize model outputs. LIME and SHAP attribute importance to features by perturbing inputs, while counterfactuals generate what-if examples through minimal, plausible changes that would flip the model’s decision \cite{wachter2017counterfactual,weilbach2024estimation}. In contrast, intrinsic approaches embed interpretability directly into models through mechanisms such as attention, prototype structures, or inherently transparent architectures \cite{rudin2019stop}. More recent developments focus on hybrid methods that integrate interpretability with fairness \cite{lofstrom2024conditional}, robustness, and causal reasoning, embedding explanation into the training pipeline \cite{cheng2025comprehensive}. Additionally, the rise of large language models has introduced capabilities to generate natural-language explanations that are context-aware and user-friendly \cite{martens2025tell}.

While these approaches have significantly advanced transparency, they typically focus on explaining why a model makes a particular decision, without explicitly communicating how reliable that decision or its explanation is. This limitation becomes especially critical in high-stakes settings, where users must assess not only plausibility but also confidence and risk \cite{yang2023explainable,vaseli2023protoasnet,ling2024improving}.

This gap has motivated increasing attention to uncertainty quantification (UQ), which we here understand as the systematic estimation and communication of uncertainty associated with model predictions and inferences. UQ is commonly distinguished into aleatoric uncertainty arising from data randomness and epistemic uncertainty resulting from limited knowledge. While uncertainty is conventionally treated probabilistically, the dominant framework in statistics and machine learning, standard approaches often fail to differentiate, or at least to communicate clearly, the respective contributions of aleatoric (irreducible) and epistemic (reducible) uncertainty \cite{hullermeier2021aleatoric}. This distinction can be exemplified as follows: a vehicle’s sales forecast might estimate an 80\% chance of exceeding the quarterly target, capturing aleatoric uncertainty due to market variability, whereas a 20\% estimate for a newly launched model may itself be uncertain because of limited historical data, reflecting epistemic uncertainty.

To ground the discussion of uncertainty representations that later appear in UAXAI methods, Hüllermeier and Waegeman \cite{hullermeier2021aleatoric} constructed a taxonomy of important machine learning approaches for representing predictive uncertainty, synthesizing how methods differ in the form of UQ they provide and in whether they explicitly separate aleatoric from epistemic components. The taxonomy contrasts, on one hand, classical probabilistic tools with Bayesian approaches (e.g., Gaussian processes, Bayesian neural networks, credal sets, reliable classification) and, on the other hand, set-valued prediction methods, such as conformal prediction and utility-based set selection, that express predictions as calibrated sets rather than points. Building on this set-valued branch, subsequent work advances conformal prediction \cite{vovk2005algorithmic} to disentangle aleatoric and epistemic uncertainty and assess their impact in practice \cite{sale2025aleatoric}.

Uncertainty-aware explainable AI (UAXAI) has emerged at the intersection of XAI and UQ, aiming to explicitly incorporate uncertainty into the explanatory process. In this survey, we define UAXAI as the class of methods that explicitly represent, communicate, or reason about uncertainty associated with model predictions, explanations, or both, with the goal of improving transparency, reliability, and human trust in AI-assisted decision making. Operationally, a method is considered UAXAI if it (i) integrates formal uncertainty estimates into explanatory outputs, (ii) uses uncertainty to qualify, validate, or modulate explanations, or (iii) explicitly distinguishes and communicates different sources of uncertainty, such as aleatoric and epistemic uncertainty, in a manner accessible to human users \cite{cohen2024uncertainty}.

From this perspective, existing UAXAI research can be broadly grouped into three complementary categories: (1) approaches that treat uncertainty itself as an explanatory signal, conveying model confidence or risk alongside predictions \cite{yapicioglu2024conformasight,lofstrom2024calibrated,lofstrom2024ce_classification,lofstrom2025calibrated}; (2) approaches that augment traditional XAI techniques with uncertainty estimates to inform or constrain explanations \cite{napolitano2024efficient,contreras2024explanation}; and (3) approaches that employ uncertainty quantification to evaluate the robustness, stability, or reliability of explanations \cite{marx2023but,narteni2023confiderai}. These categories structure the organization of the literature reviewed in this survey.

UQ and interpretability by design are deeply interconnected; indeed, UQ has been argued to provide a foundation for robust, human-centered XAI \cite{schauer2025fostering}. In this vein, recent XAI research uses UQ in two complementary ways: (i) to explain the model’s confidence in its predictions, clarifying when the model is or is not certain, and (ii) to assess the reliability and stability of the explanations produced by XAI methods. For instance, \cite{thuy2024explainability} frame uncertainty estimation as an explainability mechanism by offering local confidence cues and supporting human-in-the-loop decision-making via classification with rejection, whereas \cite{marx2023but} propose rigorous methodologies for constructing uncertainty sets to evaluate explanation consistency. In this work, we primarily focus on the former perspective, where uncertainty estimation is used to enhance transparency, strengthen decision support, and foster calibrated user trust, while also reviewing how uncertainty has been employed to assess explanation quality.

Although a growing body of research has sought to merge XAI and UQ from different perspectives, there is still no consensus on what constitutes an uncertainty-aware explanation, how different forms of uncertainty have been systematically addressed within this field, and how uncertainty-aware explanations should be evaluated in practice \cite{abeyrathna2023extension}. To address these gaps, this study undertakes a systematic survey of the literature on uncertainty-aware XAI (UAXAI), with the aims of articulating a precise and operational definition of the concept grounded in existing work, developing a structured categorization of existing contributions, and examining the evaluation methodologies proposed in prior research. In doing so, the survey not only synthesizes the state of the art but also identifies unresolved challenges, delineates current research gaps, and highlights promising directions for future developments in UAXAI.

\section{Related Work}
Although survey and review work on UQ and XAI as separate concepts is growing, only a few studies examine their convergence under uncertainty-aware explainability. One study proposed a preliminary taxonomy identifying four primary sources of uncertainty: data, AI model, XAI method, and human, highlighting the need for integrated frameworks \cite{forster2025taxonomy}. Others have framed UQ through robustness and fairness, combining these perspectives with XAI \cite{zhang2021modern}. Research has also begun exploring integration in domain-specific contexts such as medical AI \cite{salvi2025explainability} and chronic kidney disease diagnosis \cite{anoch2025uncertainty}. More recently, the focus has shifted toward operational frameworks and evaluation metrics, emphasizing trust measures to assess explanation reliability in high-stakes domains \cite{li2023scatteruq,yang2024explaining}. A dedicated review has addressed the intersection of UQ and XAI from a human perception perspective, analyzing both methodological approaches and user interpretations of uncertainty \cite{chiaburu2024uncertainty}. 

\section{Method}
To systematically identify, evaluate, and analyse studies within the scope of our research questions, we applied a systematic survey according to \cite{booth2021systematic}. The survey sought to understand how uncertainty is studied within the area of XAI. To operationalise this goal, we formulated the following sub‑questions which guided the search strategy, inclusion criteria, and subsequent coding and analysis:
\begin{enumerate}
    \item What types of uncertainty are studied, and how are they incorporated into explanations of predictive models?
    \item What are the major research directions in UAXAI?
    \item Are there any SOTA methods in UAXAI, and when did they emerge?
    \item What knowledge gaps exist within UAXAI?
    \item What future research trends can be identified?
\end{enumerate}

\subsection{Databases and Search Terms}
The study included the databases IEEE Xplore and Web of Science. Web of Science with its extensive coverage of machine learning, XAI, and Human-Computer Interaction resulted in the majority of papers. Given the multidisciplinary nature of UAXAI, this broad coverage was essential. A preliminary positioning search helped refine the terminology.

Search strings combined keywords related to uncertainty and explainability, such as "uncertainty AND explanations", "uncertainty quantification", "explainable AI", "trustworthy AI", "epistemic uncertainty", "aleatoric uncertainty", and the acronym "UAXAI" (see Appendix, Section~\ref{Search_terms}). These were tested in various combinations, using both the Author Keywords and all-fields options. For each combination, we recorded the number of hits and refined the search iteratively (e.g., adding "XAI" or "explanations" as filters). Snowball searches (backward and forward) were performed on the reference lists of relevant articles to capture additional works. For more details, see the \href{https://github.com/Moffran/UAXAI_lsurvey}{Github repository}. The initial number of papers was collected from the databases on the $5^{th}$ of April 2025 which resulted in restricting the time window to papers published until this date. The exact components and verbatim database queries for IEEE Xplore and Web of Science are provided in Appendix, Table \ref{tab:serach_strings}.

All inclusion decisions, coding, and categorizations were performed by the authors. No paper was included or excluded based solely on automated recommendations, and all extracted information was manually verified to mitigate automation bias.

As a final quality assurance, we utilised an AI-generated Scopus overview (Scopus AI) to identify potentially seminal work in uncertainty-aware XAI that may have been overlooked in the study's search strategy. We verified every representative Scopus-AI candidate (N = $25$) in our primary sources (Web of Science and IEEE Xplore) by title and/or DOI, removed duplicates, and screened with the assistance of an LLM-based audit tool; note that all inclusion/exclusion decisions were made by the authors. Artefacts are available on GitHub. This external audit contributed $0$ of $25$ additional validated studies and therefore does not alter our PRISMA counts. Artefacts (Scopus-AI report, candidate list, exact-match logs) are provided on the Github page.

\subsection{Inclusion and Exclusion Criteria}
We included peer-reviewed journal articles and conference papers that explicitly addressed uncertainty in the context of XAI, published in English. Studies were excluded if they (i) focused solely on UQ without any explainability component, (ii) presented XAI methods without discussing uncertainty, or (iii) were editorials, abstracts, theses, or grey literature. Duplicate papers were removed before screening.

\subsection{Screening and Coding Procedure}
Title and abstracts were screened independently by two reviewers against the inclusion criteria. Disagreements were resolved through discussion. Full-text screening followed to confirm relevance.

For each included paper, we recorded bibliographic details (author, title, year, venue) and applied a coding schema adapted from established qualitative practices \cite{keele2007guidelines}. The schema consisted of ten categories:
\begin{enumerate}
    \item Main focus area (XAI, UQ, UAXAI, other).
    \item Type(s) of uncertainty (epistemic, aleatoric, both, or unspecified) with notes on how uncertainty was defined.
    \item Explainability approach (post-hoc, intrinsic, hybrid, or other).
    \item Application domain (healthcare, finance, engineering, computer vision, NLP, multi-domain, or other).    
    \item User involvement (real users, simulated users, none, or not specified).
    \item Main contributions and findings (summarised, focusing on UAXAI contributions, novelty, and key results).
    \item Reported limitations (methodological or conceptual).
    \item Knowledge gaps and future directions (either stated explicitly by the authors or identified through our survey).
    \item SOTA and timeline (whether the authors define the work as SOTA within the area of UAXAI).
    \item Reviewer notes (free-form reflections of observations).
\end{enumerate}

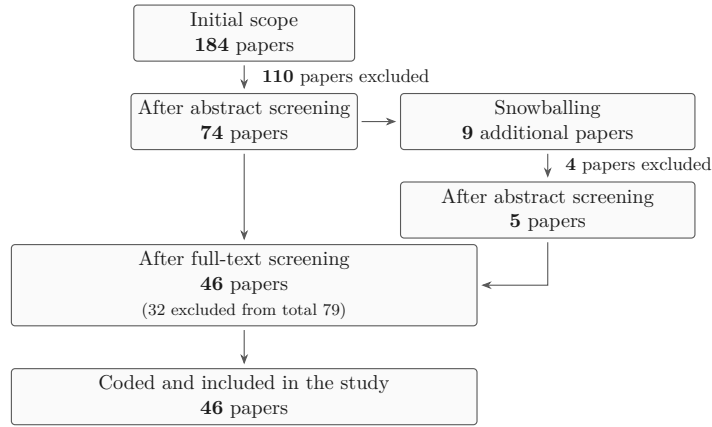
\begin{figure}[htbp!]
  \centering
  \resizebox{0.75\linewidth}{!}{
  \begin{tikzpicture}[
    node distance=6mm and 10mm, 
    >=Stealth,
    shorten >=2pt, shorten <=2pt, 
    box/.style={
      rectangle,
      rounded corners=2pt,
      draw=linegray,
      fill=fillgray,
      line width=0.6pt,
      align=center,
      minimum width=42mm, 
      minimum height=9mm,
      text=textgray
    },
    sidebox/.style={box, minimum width=56mm},
    widebox/.style={box, minimum width=72mm}, 
    lab/.style={font=\small, text=textgray}
  ]

  \node[box] (start) {Initial scope\\ \textbf{184} papers};
  \node[box, below=of start] (keep1) {After abstract screening\\ \textbf{74} papers};

  \node[sidebox, right=8mm of keep1] (snow) {Snowballing\\ \textbf{9} additional papers};
  \node[sidebox, below=of snow] (keep2) {After abstract screening\\ \textbf{5} papers};

  \node[widebox, below=18mm of keep1, minimum width=0.7\textwidth] (screen) {After full-text screening\\ \textbf{46} papers \\ \footnotesize(32 excluded from total 79)};
  \node[widebox, below=8mm of screen, minimum width=0.7\textwidth] (final) {Coded and included in the study\\ \textbf{46} papers};

 \draw[->, line width=0.6pt, draw=linegray]
    (start.south) -- node[lab, right, xshift=2mm]{\textbf{110} papers excluded} (keep1.north);

  \draw[->, line width=0.6pt, draw=linegray]
    (keep1.east) -- (snow.west);
    
 \draw[->, line width=0.6pt, draw=linegray]
    (snow.south) -- node[lab, right, xshift=2mm]{\textbf{4} papers excluded} (keep2.north);

  \draw[->, line width=0.6pt, draw=linegray]
    (keep1.south) -- (screen.north);

  \draw[->, line width=0.6pt, draw=linegray]
    (keep2.south) |- (screen.east);

  \draw[->, line width=0.6pt, draw=linegray]
    (screen.south) -- (final.north);

  \end{tikzpicture}
  }
  \caption{Selection of papers in the study.}
  \label{fig:paper_selection}
\end{figure}

\subsection{Analysis}
Following the coding of the resulting papers based on the criteria outlined in Section 3.c, a summary table encompassing ten categories was compiled and made available on the corresponding GitHub repository. The coded papers were then grouped by the generic UQ methods applied and the sources of uncertainty considered in XAI settings, accompanied by a brief summary of their main experimental contexts and organized chronologically by year. A smaller, refined summary table was subsequently created to present these grouped outcomes and highlight the main trends across years. Based on the information synthesized in the curated tables, a comprehensive decision-making pipeline was constructed to illustrate the principal sources of uncertainty and their potential accumulation across the stages of the decision-making process. 
The concept of uncertainty decomposition, along with the methodologies discussed in the reviewed papers, has been examined in detail. Subsequently, drawing on the defined sources of uncertainty, we classified their modes of incorporation into the explanations of predictive models. Finally, we outlined the SOTA approaches in UAXAI and identified the main knowledge gaps and emerging research trends that pave the way for future developments in the field.

\subsection{Limitations}
This study has collected literature, only using peer-reviewed papers from two databases (IEEE Xplore and Web of Science). Although especially Web of Science has an extensive coverage of machine learning, XAI, and Human-Computer Interaction, the limitation to only include two databases could limit the scope of the identified literature. Additionally, the snowballing method was used to find seminal papers from other databases.

To complement the manual search, we conducted en external audit using Scopus AI and an LLM-assisted screening check; however, no additional validated studies were found. 

During the search we had to omit several interesting papers, due to their preprint status which could perhaps given us a clearer picture of the area. However, it is always a risk that interesting papers is published after the limitation of the time window.

\section{Results}

The selection of papers followed a three-phase strategy, as viewed in Figure \ref{fig:paper_selection}: i) Abstract reading of initial scope of papers, ii) snowballing and abstract reading of additional interesting papers iii) Full-text screening of remaining included papers. The initial number of unique papers in the corpus, based on the combination of search terms, was 184. 

\begin{table}[h!]
    \centering
    \footnotesize
    \caption{Full-text exclusion reasons (n=32).}
    \begin{tabular}{L{7cm} L{3cm}}
    \hline
    \textbf{Description} & \textbf{no. of papers} \\
    \hline
    No uncertainty quantification & 13  \\
    Not XAI-related               & 13  \\
    No explanation component      & 5  \\
    No access                     & 1  \\
    \hline
    Total & 32 \\
    \end{tabular}
    \label{tab:paper_exclusion}
\end{table}

Abstract reading resulted in the exclusion of 110 papers, and 74 papers were read and coded according to instructions on the GitHub page. The snowballing identified 9 additional papers, among which 4 were excluded in the abstract reading. The remaining 5 were included in the corpus, resulting in 79 papers read in full. For borderline cases at the full-text screening stage, papers were included only if they explicitly integrated quantified uncertainty into the explanation process itself, rather than merely reporting predictive uncertainty or using uncertainty for internal model optimisation. When uncertainty was reported solely as a performance or calibration property without influencing the explanation output or its interpretation, the paper was excluded. From the 79 papers, 32 papers where omitted due to mainly two reasons, see Table \ref{tab:paper_exclusion}: either not including an uncertainty estimation method (14 papers), or covering either XAI or UQ but not UAXAI (13 papers), resulting in 46 papers. All analyses in the study are built on the results from coding the resulting 46 papers. 

{
\centering
\footnotesize
\setlength{\tabcolsep}{3pt} 
\begin{longtable}{C{0.8cm} L{7cm} L{1.5cm} L{2.5cm}}
\caption{Overview of included papers on uncertainty and explainable AI.}
\label{tab:uncertainty-xai-papers}\\
\toprule
\textbf{Year} & \textbf{Paper title} & \textbf{Type(s) of Uncertainty Studied} & \textbf{Uncertainty Handling Methods}\\
\midrule
\endfirsthead
\toprule
\textbf{Year} & \textbf{Paper title} & \textbf{Type(s) of Uncertainty Studied} & \textbf{Uncertainty Handling Methods}\\
\midrule
\endhead

2020 & General Pitfalls of Model-Agnostic Interpretation Methods for Machine Learning Models \cite{molnar2020general}  & Unclear or unspecified & Narrative \\

2021 & BayLIME: Bayesian Local Interpretable Model-Agnostic Explanations \cite{zhao2021baylime} &  Epistemic & Model; Quantitative \\
2021 & Generating Interpretable Counterfactual Explanations By Implicit Minimisation of Epistemic and Aleatoric Uncertainties \cite{schut2021generating} & Both & Model; Predictive; Quantitative  \\
2021 & Reliable Post hoc Explanations: Modelling Uncertainty in Explainability \cite{slack2021reliable}  & Unclear or unspecified & Model; Quantitative \\
2021 & Uncertainty Class Activation Map (U-CAM) Using Gradient Certainty Method \cite{patro2021uncertainty} & Both & Predictive; Model; Visual; Quantitative  \\
2021 & Uncertainty interpretation of the Machine Learning Survival Model Prediction \cite{utkin2021uncertainty} & Aleatoric & Predictive; Quantitative  \\

2022 & Select Wisely and Explain: Active Learning and Probabilistic Local Post-hoc Explainability \cite{saini2022select} & Both & Predictive; Model; Quantitative   \\

2023 & But Are You Sure? An Uncertainty-Aware Perspective on Explainable AI \cite{marx2023but}  & Epistemic & Model; Narrative  \\
2023 & CONFIDERAI: CONFormal Interpretable-by-Design score function for Explainable and Reliable Artificial Intelligence \cite{narteni2023confiderai} & Unclear or unspecified & Predictive; Quantitative  \\
2023 & Distributional Prototypical Methods for Reliable Explanation Space Construction \cite{joo2023distributional} & Both & Predictive; Model; Visual; Quantitative \\
2023 & Explainability meets uncertainty quantification: Insights from feature-based model fusion on multimodal time series \cite{folgado2023explainability} & Both & Predictive; Model; Quantitative \\
2023 & Explainable Uncertainty Quantifications For Deep Learning-based Molecular Property Prediction \cite{yang2023explainable} & Both & Model; Visual; Quantitative\\
2023 & Explaining Prediction Uncertainty in Text Classification: The DUX Approach \cite{andersen2023explaining} & Epistemic & Model; Quantitative \\
2023 & Extension of Regression Tsetlin Machine for Interpretable Uncertainty Assessment \cite{abeyrathna2023extension} & Both & Predictive; Quantitative; Narrative  \\
2023 & Generative Perturbations Analysis for Probabilistic Black-Box Anomaly Attribution \cite{ide2023generative} & Epistemic & Visual; Model \\
2023 & Joint Gaussian mixture model for versatile deep visual model explanation \cite{xie2023joint} & Unclear or unspecified & Predictive; Quantitative; Narrative\\
2023 & ProtoASNet: Dynamic Prototypes for Inherently Interpretable and Uncertainty-Aware Aortic Stenosis Classification in Echocardiography \cite{vaseli2023protoasnet} & Aleatoric & Quantitative\\
2023 & ScatterUQ: Interactive Uncertainty Visualizations for Multiclass Deep Learning Problems \cite{li2023scatteruq} & Epistemic & Predictive; Visual; Quantitative\\

2024 & A framework for counterfactual explanation of predictive uncertainty in multimodal models \cite{qiu2024framework} & Epistemic & Model; Quantitative \\
2024 & A Trustworthy Counterfactual Explanation Method With Latent Space Smoothing \cite{li2024trustworthy} & Unclear or unspecified  & Quantitative; Other\\
2024 & Addressing label noise in leukemia image classification using small loss approach and pLOF with weighted-average ensemble \cite{aziz2024addressing} & Epistemic & Model; Visual \\
2024 & Application of spatial uncertainty predictor in CNN-BiLSTM model using coronary artery disease ECG signals \cite{seoni2024application} & Both & Model; Visual; Quantitative \\
2024 & Boundary-Aware Uncertainty for Feature Attribution Explainers \cite{hill2024boundary} & Both & Model; Visual; Quantitative \\
2024 & Calibrated Explanations for Multi-Class \cite{lofstrom2024calibrated} & Both & Predictive; Visual; Quantitative \\
2024 & Calibrated explanations: With uncertainty information and counterfactuals \cite{lofstrom2024ce_classification} & Both & Predictive; Quantitative\\
2024 & Communicating Uncertainty in Machine Learning Explanations: A Visualization Analytics Approach for Predictive Process Monitoring \cite{mehdiyev2024communicating} & Epistemic Uncertainty & Model; Visual; Quantitative \\
2024 & Conditional Calibrated Explanations: Finding a Path Between Bias and Uncertainty \cite{lofstrom2024conditional} & Both & Predictive; Visual; Quantitative \\
2024 & ConformaSight: Conformal Prediction-Based Global and Model-Agnostic Explainability Framework \cite{yapicioglu2024conformasight} & Unclear or unspecified & Model; Quantitative \\
2024 & DUE: Dynamic Uncertainty-Aware Explanation Supervision via 3D Imputation  \cite{zhao2024due}& Both & Predictive; Model; Visual; Quantitative\\
2024 & Efficient Neural Network-Based Estimation of Interval Shapley Values \cite{napolitano2024efficient} & Unclear or unspecified & Model; Quantitative\\
2024 & Estimation of Counterfactual Interventions under Uncertainties \cite{weilbach2024estimation} & Epistemic & Model; Quantitative\\
2024 & Explanation of Deep Learning Models via Logic Rules Enhanced by Embeddings Analysis, and Probabilistic Models \cite{contreras2024explanation} & Aleatoric & Model; Narrative \\
2024 & Explaining Probabilistic Bayesian Neural Networks for Cybersecurity Intrusion Detection \cite{yang2024explaining} & Both & Model; Predictive\\
2024 & Improving Explainable Object-induced Model through Uncertainty for Automated Vehicles \cite{ling2024improving} & Epistemic & Predictive; Quantitative  \\
2024 & Investigating the Impact of Model Instability on Explanations and Uncertainty \cite{marjanovic2024investigating} & Epistemic & Predictive; Quantitative \\
2024 & Quantifying uncertainty in graph neural network explanations \cite{jiang2024quantifying} & Both & Model; Quantitative  \\
2024 & Towards Modelling Uncertainties of Self-Explaining Neural Networks via Conformal Prediction \cite{qian2024towards} & Unclear or unspecified & Predictive; Quantitative \\
2024 & ULTRA-AIR UltraSound Landmark tracking for Real-time Anatomical Airway Identification and Reliability Check \cite{khodagholi2024ultra} & Both & Model; Quantitative  \\
2024 & Uncertainty in XAI: Human Perception and Modelling Approaches \cite{chiaburu2024uncertainty} & Unclear or unspecified  & Predictive; Model; Visual; Quantitative; Narrative\\
2024 & Uncertainty-Aware Explainable AI as a Foundational Paradigm for digital twins \cite{cohen2024uncertainty} & Unclear or unspecified & Narrative   \\

2025 & An explainable Liquid Neural Network combined with path aggregation residual network for an accurate brain tumour diagnosis \cite{shaheema2025explainable} & Aleatoric & Visual; Quantitative\\
2025 & Augmenting post-hoc explanations for predictive process monitoring with uncertainty quantification via conformalized Monte Carlo dropout \cite{mehdiyev2025augmenting} & Epistemic & Predictive; Model; Visual; Quantitative\\
2025 & Calibrated explanations for regression \cite{lofstrom2025calibrated} & Both & Predictive; Quantitative\\
2025 & Explainability and uncertainty: Two sides of the same coin for enhancing the interpretability of deep learning models in healthcare \cite{salvi2025explainability} & Both & Narrative\\
2025 & Explaining predictive uncertainty by exposing second-order effects \cite{bley2025explaining}& Epistemic & Predictive; Model; Visual; Quantitative  \\
2025 & Integrating permutation feature importance with conformal prediction for robust Explainable Artificial intelligence in predictive process monitoring \cite{mehdiyev2025integrating} & Both & Predictive; Quantitative\\

\bottomrule
\end{longtable}
}

\subsection{Definitions and Understandings of Uncertainty in XAI}
Building on the operational definition of uncertainty-aware explainable artificial intelligence (UAXAI) introduced in Section~\ref{introduction}, this section examines how uncertainty has been conceptualized and categorized within the literature on uncertainty-aware explainable AI. Uncertainty has long been regarded as an inherent feature of learning systems, but its role within XAI has only recently begun to receive systematic attention. Rather than viewing uncertainty merely as a nuisance, Hüllermeier and Waegeman \cite{hullermeier2021aleatoric} emphasize that it is a multifaceted concept that can be classified, measured, and meaningfully communicated. Building on this perspective, recent studies have broadened the scope by conceptualizing uncertainty as spanning the entire socio-technical pipeline. In this context, Förster et al. \cite{forster2025taxonomy} identify four complementary sources: \textit{data uncertainty}, arising from noise, bias, or incompleteness; \textit{model uncertainty}, linked to assumptions, architecture, or training; \textit{method uncertainty}, reflecting instability or disagreement among explanation techniques; and \textit{human uncertainty}, capturing variability in how individuals interpret and act upon explanations. Within this broader landscape, Förster et al. characterize UAXAI as explanation methods designed to explicitly expose and communicate multiple forms of uncertainty along the socio-technical pipeline.

\begin{figure}[h!]
    \centering
    \includegraphics[width=0.9\linewidth]{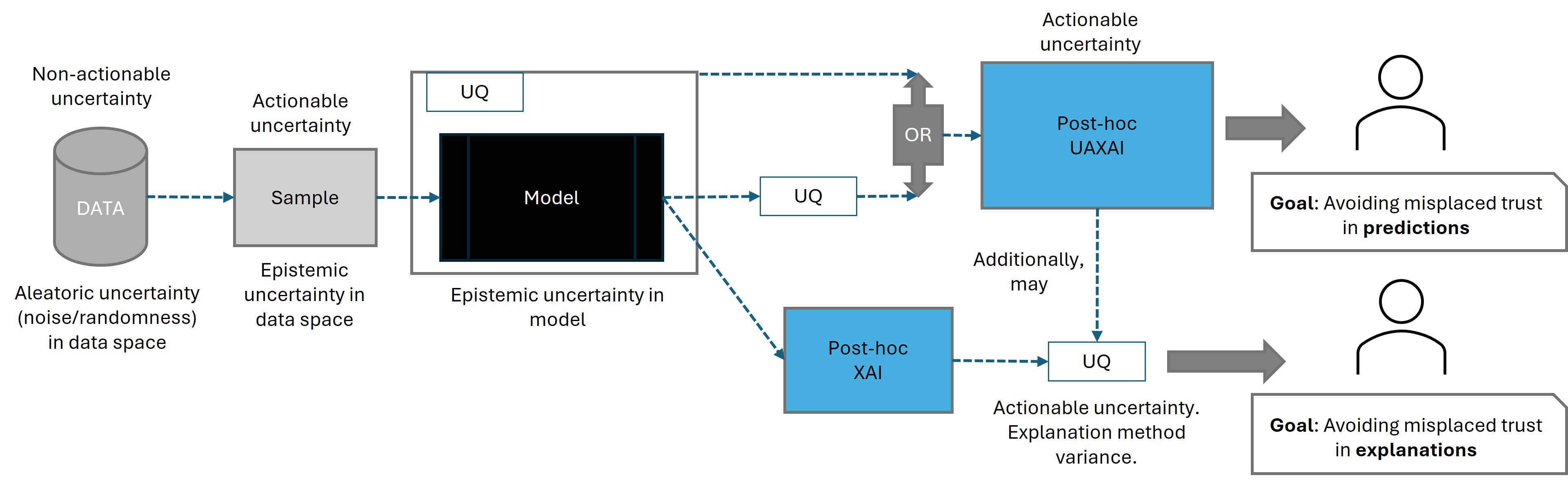}
    \caption{Illustration of sources of uncertainty across the technical stages of the decision pipeline; from data collection to explanation methods. The figure emphasizes technical sources only, user-related uncertainties are not included.}
    \label{fig:pipeline}
\end{figure}

The problematization of what constitutes an explanation with uncertainty remains unresolved. For instance, it is unclear whether a visualization of a feature value together with its variance should be considered an uncertainty-aware explanation or whether it merely represents descriptive statistical dispersion. This ambiguity points to a deeper issue concerning how uncertainty itself is defined. The vagueness observed in parts of the literature may be attributed to the multiple sources of uncertainty that can arise along the decision-making pipeline, as illustrated in Figure \ref{fig:pipeline}. Early contributions often lacked a nuanced treatment of uncertainty, relying on vague or inconsistent terminology. Some included studies conflate variance with uncertainty, while others adhere to more rigorous formulations, such as those proposed by Hüllermeier \cite{hullermeier2021aleatoric}, which clearly distinguish between aleatoric and epistemic uncertainty. From around 2023 onward, definitions generally became more precise, and the distinction between aleatoric and epistemic uncertainty was more explicitly introduced. As interest in the topic increased, taxonomies and surveys began to emerge, several of which address the entire socio-technical lifecycle from data collection to user decision-making. One such taxonomy emphasizes the risk of uncertainty accumulation across the pipeline, underscoring the importance of reducing uncertainty wherever possible to prevent compounding effects \cite{blair2024transfer}.

To address the complexity arising from the divergence of uncertainty sources, the taxonomy proposed in \cite{forster2025taxonomy} identifies four primary sources of uncertainty, further differentiated into 38 subcategories ranging from technical aspects such as data noise to human-centred factors such as users’ cognitive limitations. In the present study, however, the scope is deliberately narrower and aligned with the operational definition of UAXAI adopted in this survey: the focus lies on XAI methods that explicitly incorporate uncertainty information into the explanation itself. By restricting the analysis in this way, the study highlights approaches that not only expose the internal reasoning of models but also make uncertainty an integral component of the explanatory output.

Within this context, aleatoric uncertainty is typically defined as noise or randomness inherent in the data. Two variants are commonly distinguished: \textit{heteroscedastic} uncertainty, where the degree of noise varies across the input space, and \textit{homoscedastic} uncertainty, where it remains constant \cite{kendall2017uncertainties}. This distinction highlights the importance of sampling strategies in capturing the underlying data distribution. Although aleatoric uncertainty is notoriously difficult to quantify, several studies attempt to approximate it \cite{sale2025aleatoric,zhang2024one}. While noise in real-world data cannot be eliminated, awareness of sampling design, algorithmic choices, and transparency in model decisions remains crucial when striving to reduce uncertainty accumulation.

A second dimension is often expressed through the distinction between \textit{global} and \textit{local uncertainty}. Global uncertainty refers to the reliability of the model as a whole, whereas local uncertainty is associated with the trustworthiness of individual predictions or explanations. The latter is typically emphasized in XAI, as user decision-making is directly influenced by local explanations \cite{lofstrom2024ce_classification}. 

Uncertainty has also been conceptualized as variance between explanation methods, whereby divergence across saliency maps, counterfactuals, or attribution scores is interpreted as an indication of instability in the explanatory process itself. This perspective has been referred to as \textit{explainer} or \textit{explanation uncertainty} \cite{chiaburu2024uncertainty}. The concept is closely linked to the robustness and stability of post-hoc XAI methods \cite{slack2021reliable}, with the aim of avoiding misplaced trust in explanations (see Figure \ref{fig:pipeline}).

\begin{table}[h!]
    \centering
    \footnotesize
    \caption{Origins of uncertainty in the technical aspects of the socio-technical ML pipeline, their nature, and possible mitigation strategies.}
    \begin{tabular}{L{1.7cm} L{3cm} C{1.2cm} L{1.9cm} L{3cm}}
    \hline
    \textbf{Origin} & \textbf{Pipeline stage} & \textbf{Type} & \textbf{Mitigation} & \textbf{Uncertainty mitigation methods} \\
    \hline
    Intrinsic data variability & Real-world randomness, measurement error & Aleatoric & Not reducible & -- \\
    Data sampling & Training, validation, and test set design & Epistemic & Improved sampling; calibration split & Calibration set \\
    Model specification & Algorithm, architecture, hyperparameters & Epistemic & Well-calibrated models and post-hoc calibration & \textbf{Models:} Random Forest, XGBoost, neural networks. \newline \textbf{Calibration:} Conformal, Platt, Isotonic \\
    Post-model & Post-hoc interpretability methods & Epistemic & Robust and stable UAXAI & UAXAI methods. \\
    \hline
    \end{tabular}
    \label{tab:uncertainty-origins}
\end{table}

Table \ref{tab:uncertainty-origins} summarizes the possible origins of uncertainty along the decision pipeline. Some sources are actionable and can be mitigated through methodological choices, while others are non-actionable and reflect intrinsic properties of the data or task. For example, real-world data inevitably exhibits inherent variability. However, uncertainty introduced at the sampling stage may be reduced or amplified depending on the chosen sampling strategy. Although sampling alone cannot eliminate uncertainty, the use of a calibration set can align the trained model with the underlying data distribution, thereby reducing discrepancies between predicted and observed outcomes.

Similarly, a UAXAI method does not directly reduce uncertainty within the system itself but instead supports users in making more informed decisions, potentially reducing user uncertainty. Finally, ensuring that explanation methods are robust and stable helps to minimize explanation variance, preventing additional uncertainty from being introduced by the explanatory process itself.

\subsection{Uncertainty Quantification Trends and Patterns Over Time}
Early works treated UQ and XAI largely as separate research threads. However, by 2023--2025, the literature shows a marked convergence toward UAXAI.  The integration proceeds along three distinct directions:  

\begin{itemize}  
    \item \textbf{Use UQ to select or weight explanations} -- for example, Gaussian Process uncertainty is combined with active sampling to stabilize post-hoc explanations \cite{saini2022select}; multimodal time-series fusion weights feature attributions according to model uncertainty \cite{folgado2023explainability}; and ensemble models in medical imaging apply uncertainty-weighted averaging \cite{aziz2024addressing}.  
    \item \textbf{Use UQ to train or constrain explanations} -- for example, predictive uncertainty is used to supervise explanation quality during training in \cite{zhao2024due}, while distributional prototype methods incorporate uncertainty into the latent representation \cite{li2024trustworthy} to yield more reliable explanation spaces \cite{joo2023distributional}.  
    \item \textbf{Explain the uncertainty itself} -- for example, calibration is used to inform factual and counterfactual explanations to explicitly communicate prediction intervals alongside feature importances \cite{lofstrom2024ce_classification}; predictive uncertainty is used to gain higher-order feature interactions \cite{bley2025explaining}; and aleatoric and epistemic contributions is decomposed in saliency maps \cite{jiang2024quantifying}. 
\end{itemize}  

\begin{figure}[h!]
    \centering
    \includegraphics[width=0.85\linewidth]{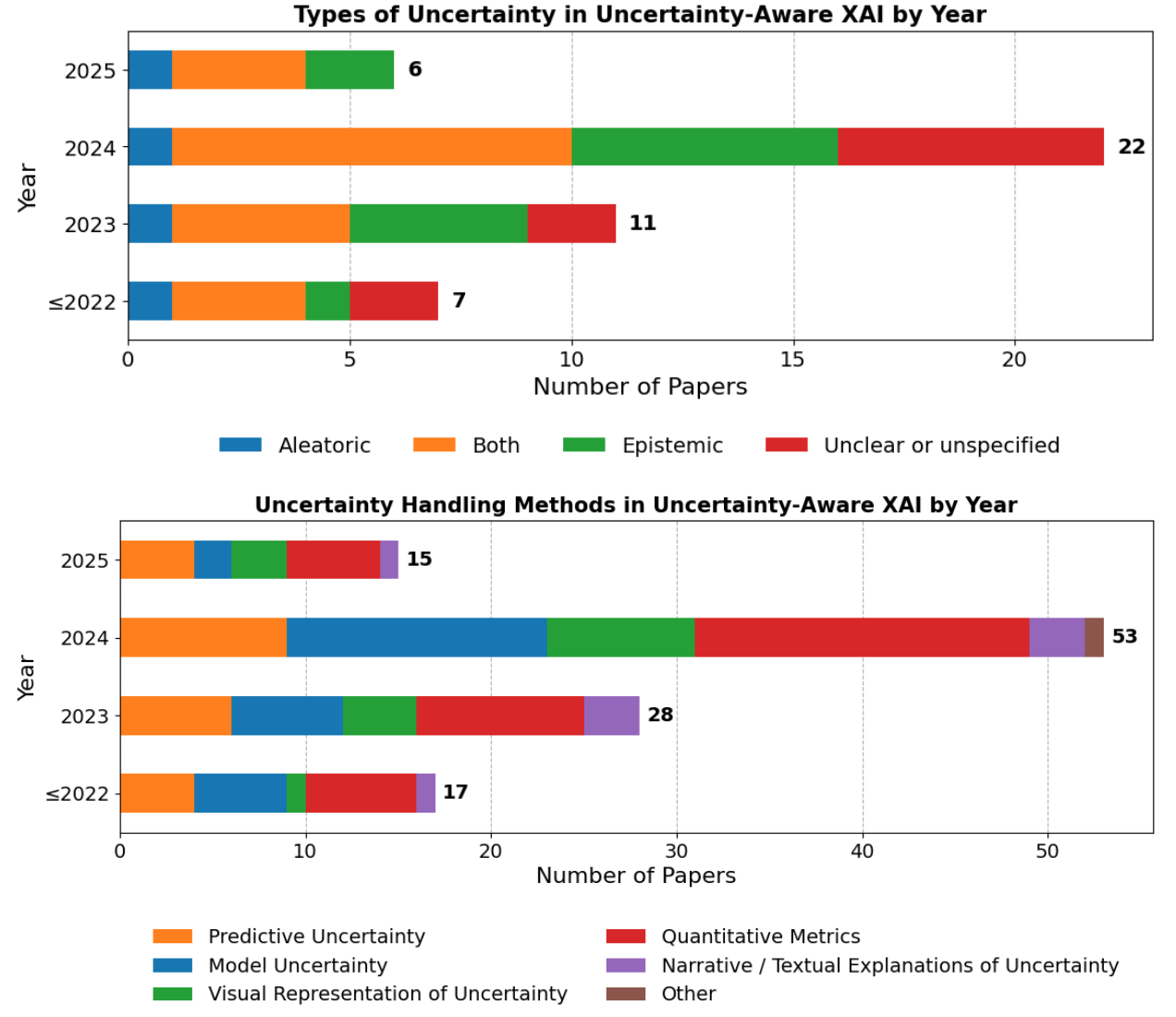}
    \caption{Evolution of uncertainty types and handling methods in Uncertainty-Aware XAI literature. The years 2021 and 2022 were merged (<=2022) due to limited publications. Several uncertainty handling methods were sometimes included in the same paper, accounting for the total sum being larger than 46 in the lower plot.}
    \label{fig:evolution}
\end{figure}

Recent trends in UQ for XAI, summarized in Figure~\ref{fig:evolution}, show a clear shift from foundational methods toward more hybrid and application-oriented approaches. This early phase was marked by Bayesian surrogate explanations and uncertainty maps \cite{zhao2021baylime, patro2021uncertainty}, which extended the toolbox and enabled a more nuanced decomposition of aleatoric and epistemic uncertainty. Building on this foundation, \textit{Bayesian} approaches have continued to dominate the field, capturing epistemic uncertainty through priors, posteriors, and ensemble formulations till 2023 \cite{saini2022select} (17 of 47 papers). Alongside these, \textit{Monte Carlo dropout} emerged as a lightweight baseline, valued for its ease of integration and relatively low computational cost \cite{mehdiyev2025augmenting, andersen2023explaining} (6 out of 47 papers), while calibration methods maintained their appeal as simple, model-agnostic techniques for improving reliability \cite{utkin2021uncertainty} (7 out of 47 papers).

From 2024 onwards, we observe calibration-focused methods (8 of 28 papers) becoming more prominent in our corpus, offering distribution-free guarantees and well-calibrated intervals that align with practical decision-making needs \cite{lofstrom2024ce_classification, qian2024towards}. By 2025, hybrid methods extend this line of work, with conformalized Monte Carlo dropout \cite{mehdiyev2025augmenting} and second-order explanations \cite{bley2025explaining} explicitly integrating UQ with interpretability. At the same time, explanation method variance is gaining recognition as a theme, emphasizing that post-hoc explainers themselves introduce uncertainty, not only the predictive models they interpret \cite{joo2023distributional, bley2025explaining, seoni2024application}. Collectively, these developments highlight a shift from producing uncertainty estimates to embedding UQ into the very fabric of explanations, clarifying both the content and the confidence of model outputs.

\subsection{Decomposition of Uncertainty} 
Uncertainty decomposition remains an open and context-dependent concept without a unified definition. While the distinction between aleatoric and epistemic uncertainty is widely referenced \cite{hullermeier2021aleatoric,der2009aleatory}, these categories often overlap and depend on the specific modeling context. Across the investigated articles, uncertainty decomposition is generally approached by separating or attributing different sources of uncertainty within AI models to enhance interpretability and trust. 

Types of uncertainty decomposition and their descriptions:
\begin{itemize}
\item \textbf{Aleatoric vs. Epistemic Uncertainty Decomposition}: separates uncertainty caused by data noise or randomness (aleatoric) from that due to limited model knowledge or parameters (epistemic). 
\item \textbf{Data / Model / Feature-Level Decomposition}: Attributes overall predictive uncertainty to specific sources such as input data quality, model variability, or influential features and modalities. 
\item \textbf{Interaction-Based Uncertainty Decomposition}: Explains uncertainty emerging from interactions among features, capturing second-order effects beyond individual feature contributions. 
\item \textbf{Human–Technical (Conceptual) Decomposition}: Considers both computational and perceptual aspects of uncertainty, linking model and data uncertainty to human understanding and trust. 
\label{tab:uncertainty_decomposition}
\end{itemize}

Though most studies distinguish between aleatoric and epistemic uncertainty as the primary form of decomposition \cite{salvi2025explainability}, the literature demonstrates several complementary perspectives, summarized above. Some works extend the classical view by distributing uncertainty across data, model \cite{jiang2024quantifying}, and feature levels \cite{zhang2025interpret,reddy2024uncertainty}, while others analyze interaction-based effects \cite{bley2025explaining}, where uncertainty arises from feature interdependencies rather than isolated inputs. In multimodal or hierarchical models, decomposition is performed across modalities or layers, identifying which components contribute most to predictive uncertainty \cite{qiu2024framework}. A few studies also integrate human–technical perspectives, framing decomposition as both a computational and perceptual process that connects model-derived uncertainty with human trust and interpretability \cite{molnar2020general,prabhudesai2023understanding}. Overall, uncertainty decomposition serves as a practical abstraction rather than a strict theoretical framework, with precise quantification of each uncertainty type remaining an open challenge.

\subsection{Evaluation Metrics} 
Across the reviewed literature, evaluation practices in UAXAI remain highly heterogeneous and largely model-centred rather than explanation-centred. Most studies assess explanations indirectly, through model performance or quantitative stability analyses, rather than by systematic user evaluation. Only a few exceptions, typically within healthcare or human-computer-interaction (HCI) contexts, explicitly consider human interpretability or trust \cite{molnar2020general}. When included, human assessments are often qualitative or expert-based, focusing on whether explanations align with domain knowledge.\\

Common approaches for evaluating explanations quantitatively include:
\begin{itemize}
\item \textbf{Model-accuracy metrics} (e.g., RMSE, accuracy, F1-score), although these risk conflating model performance with explanation quality \cite{mehdiyev2025integrating,khodagholi2024ultra}.
\item \textbf{Robustness and stability metrics}, measuring consistency under perturbations, resampling, or retraining (e.g., variance in feature importance, Jaccard similarity, local fidelity) \cite{ide2023generative,marjanovic2024investigating}. Risks conflating UAXAI with uncertainty within the XAI method.
\item \textbf{Uncertainty-specific measures}, such as coverage probability, calibration error, entropy, interval width, or mutual information \cite{lofstrom2024conditional,yapicioglu2025conformasegment}.
\item \textbf{Faithfulness and fidelity}, comparing explanation attributions to the model’s actual decision process.
\item \textbf{Comprehensibility and complexity}, often gauged via sparsity or cognitive load.
\end{itemize}


Traditional XAI evaluation focuses on \textit{faithfulness} and \textit{robustness}, whereas UAXAI adds the complementary dimensions of \textit{reliability} and \textit{confidence}. In other words, a standard explanation is judged by what it shows, while a UAXAI explanation is also judged by how certain or stable that information is. This shift introduces new evaluation objectives: rather than merely describing model decisions, UAXAI methods aim to clarify when and why explanations can be trusted. Consequently, evaluation metrics must capture quantities such as \textit{variance}, \textit{coverage}, and \textit{consistency}, aspects typically overlooked by conventional XAI metrics.

Among existing explanation categories, counterfactual explanations are particularly compatible with this perspective. Both are generative by nature, exploring how predictions change under perturbation and focusing on local uncertainty reduction, that is, identifying minimal feature changes that alter predictions while limiting epistemic ambiguity.

Despite these advances, key gaps remain. Existing metrics rarely capture the epistemic–aleatoric decomposition, the propagation of uncertainty from model to explanation, or the calibration of user trust. A lack of proper uncertainty calibration can have significant societal implications, such as wrongful loan denials, which highlight the risk of miscalibrated trust. A unified evaluation framework is therefore needed, one that jointly measures confidence calibration, robustness under perturbation, and the impact on human decisions. Counterfactual-based and conformal approaches appear especially promising for combining interpretability with reliability.

\subsection{Defining State-of-the-Art in Uncertainty-Aware XAI}
Defining SOTA is difficult because techniques span model-specific and model-agnostic regimes. The common denominator is uncertainty that is computed, communicated, and validated as part of the explanation. Our results indicate three recurring levers (with some examples in the material): use UQ to weight or select explanations \cite{zhao2024due,saini2022select,patro2021uncertainty}, use UQ to constrain learning of explanations \cite{joo2023distributional,hill2024boundary}, and explain the uncertainty itself \cite{zhao2021baylime}. SOTA implementations increasingly embed these levers inside calibrated or conformal frameworks, and they make decomposition explicit when it helps users reason about risk. This responds to two gaps: heterogeneous evaluation that often conflates model performance with explanation quality, and weak attention to how uncertainty propagates from model to explainer. SOTA closes these by reporting coverage, calibration, and stability alongside faithfulness, and by analyzing explanation variance. Counterfactual explanations align well because they test local changes and can reduce epistemic ambiguity; conformal approaches add distribution-free guarantees. The field remains fast-moving, so we frame SOTA as practices with evidence rather than a canon of methods, i.e., SOTA methods tell the user what the model says and how sure that statement is, with evidence.

\section{Discussion}
The XAI literature encompasses a vast range of methods, including both model specific and model agnostic explanations, which makes it challenging to compare and contrast the various methods. The same challenge arises within the field of UAXAI, as some uncertainty-aware explanations rely heavily on specific model architectures and/or explanation techniques, while others are designed for a broader class of model agnostic XAI methods. As a result, defining a clear SOTA in the UAXAI literature remains difficult. However, the incorporation of UQ into the explanations might itself be regarded as a SOTA development of XAI, taking a step towards a more risk-aware and nuanced view of what a good explanation of an AI algorithm should entail. 

The survey finds that the majority of studies overlook the human perception and usability of the proposed explanations. While the primary objective of UAXAI is to generate explanations that incorporate and communicate different types of uncertainty in the underlying model and data, the broader purpose of XAI remains to enhance human understanding and trust in AI systems. The cognitive dimension of how humans interpret the UAXAI methods, in contrast conventional XAI methods that disregard uncertainty, is a promising topic for further research. 

Another important trend in the literature is the decomposition of uncertainty into aleatoric and epistemic, as pioneered by \cite{hullermeier2021aleatoric}. Many of the reviewed UAXAI methods incorporate such a decomposition, either explicitly through mathematical derivations or on a more conceptual level. An important next step for UAXAI research concerns the practical implication of these decompositions for the user's interpretations of the explanations. For instance, should users place less confidence in explanations that exhibit high epistemic uncertainty, suggesting that the model fails to capture underlying patterns adequately? Or should they be more cautious when explanations reflect predominantly aleatoric uncertainty, indicating inherent noise in the data? Addressing how the different types of uncertainty should be interpreted remains a key challenge for the adoption of uncertainty-aware explanations. 

A final methodological challenge encountered in the literature survey concerns the fast moving nature of the UAXAI field. Several promising and relevant papers had to be excluded due to their preprint status at the time of writing. Several relevant papers have also been published after the breaking date for our data collection, further highlighting the momentum of the research area, and we anticipate that research in this area will continue to evolve in the coming years. 

\section{Conclusion and Suggestions of Future Work} 
Uncertainty-aware explanations have emerged as an essential component of XAI. While the influence of model uncertainty on decision-making has been acknowledged in machine learning for some time, the systematic integration of uncertainty into explanation methods remains in its early stages. Early studies addressed uncertainty only in general terms, whereas more recent research distinguishes between aleatoric and epistemic uncertainty and examines their respective roles in shaping trustworthy explanations.

In current XAI research, uncertainty is most commonly quantified using three methodological categories, each with distinct trade-offs in mathematical rigour, computational costs, and coverage guarantees. These include Bayesian methods, Monte Carlo dropout techniques, and conformal prediction-based approaches, the latter of which provides finite-sample coverage guarantees. Furthermore, three different approaches can be identified in how uncertainty is incorporated into explanation pipelines: i) using uncertainty to assess the trustworthiness of predictions, ii) using it to enhance training data for improved model accuracy, and iii) explicitly explaining uncertainty itself as part of the interpretive output.

Over time, the methodological landscape has evolved from primarily Bayesian and variance-based UQ towards the integration of conformal prediction and calibrated interval estimates, approaches that combine interpretability with formal guarantees. Taken together, these developments suggest a field that is moving beyond the mere estimation of uncertainty toward embedding it as a structural element of explainability, with an increasing emphasis on calibration, reliability, and user-centred communication of predictive confidence.\\ \\ \\

Our study suggests four directions on which UAXAI should prioritise in future research:
\begin{itemize}
    \item \textbf{Extending formal uncertainty guarantees}, such as those from conformal prediction and credal sets, to combine interpretability with formal guarantees.
    \item  \textbf{Standardised evaluation protocols} to assess predictive reliability, explanation stability, and user interpretability.
    \item  \textbf{User studies} to examine how different representations of uncertainty influence trust calibration and decision quality.
    \item \textbf{Uncertainty decomposition}, providing insights into the cause of uncertainty in terms of aleatoric and epistemic uncertainty. 
\end{itemize}



\bibliography{sample} 

\appendix \label{A}
\section{Additional methodological details}
\subsection{Search Terms} \label{Search_terms}
The main search terms in the study was the following:
\begin{itemize}
    \item Uncertainty + Explanations + XAI
    \item UAXAI
    \item Explainable+ AI
    \item Uncertainty + Quantification
    \item Explainable + Artificial + Intelligence
    \item Trustworthy + AI
    \item Epistemic uncertainty
    \item Epistemic and aleatoric uncertainty
    \item Aleatoric uncertainty
\end{itemize}

\subsection{Combination of Search Terms, including overlap}
The search terms was combined together, see table \ref{tab:serach_strings}, and used to extract relevant papers from the databases. The exact verbatim that were used correspond to the first column in the table. The total number of papers in the table are higher than the number in the initial corpus, since several of the searches resulted in redundant information. The search engines in the databases automatically removed duplicates when adding the search results to a saved list which resulted in a total of $184$ papers.

\begin{table}[h!]
    \centering
    \footnotesize
    \caption{Combination of search terms run on IEE Xplore and Web of Science the $5^th$ of April 2025, including overlap.}
    \begin{tabular}{L{4cm} L{2cm} L{2cm} L{2cm} L{1cm}}
    \hline
    \textbf{Search string} & \textbf{Section of Paper} & \textbf{Research Area(s)} & \textbf{Refined by} & \textbf{No. of papers}\\
    \hline
    Uncertainty + explanations & Author keywords & Computer Science & - & 33 \\
    Uncertainty + explanations + XAI & Author keywords & Computer Science & - & 1 \\
    UAXAI & Author keywords & Computer Science & - & 0 \\
    UAXAI & all fields & - & -& 1 \\
    Trustworthy + AI & Author keywords & Computer Science & Uncertainty & 27\\
    Trustworthy + AI & Author keywords & Computer Science & Uncertainty and Explanations & 12\\
    Epistemic uncertainty & Author keywords & Computer Science & explanations & 2\\
    Uncertainty-aware + explanations & All fields & - & XAI & 2\\
    epistemic and aleatoric uncertainty & Author keywords &- & - & 0\\
    epistemic and aleatoric uncertainty & All fields & Computer Science & explanations & 5\\
    Explainable + AI & Author keywords & Computer Science & Uncertainty & 78 \\
    Explainable – Artificial – Intelligence & Author keywords & Computer Science &  Uncertainty & 71\\
    Uncertainty + quantification & Author keywords & Computer Science & Explanations & 17\\
    Uncertainty + quantification & All fields & Computer Science & Explanations & 32\\
    \hline
    \multicolumn{4}{L{4 cm}}{Total} & 281 \\
    \end{tabular}
    \label{tab:serach_strings}
\end{table}

\end{document}